\documentclass{article}

\PassOptionsToPackage{numbers, sort&compress}{natbib}

\usepackage[preprint]{neurips_2024}

\usepackage[utf8]{inputenc} 
\usepackage[T1]{fontenc}    
\usepackage{hyperref}       
\usepackage{url}            
\usepackage{booktabs}       
\usepackage{amsfonts}       
\usepackage{nicefrac}       
\usepackage{microtype}      
\usepackage{xcolor}         

\usepackage{microtype}
\usepackage{amsmath}
\usepackage{amssymb}       
\usepackage{numprint}
\usepackage{xspace}
\usepackage{multirow}
\usepackage{multicol}

\newcommand{\first}[1]{{\color{red}{#1}}}
\newcommand{\second}[1]{{\color{blue}{#1}}}

\newcommand{\etal}{et al. }
\newcommand{\ie}{\emph{i.e}. }
\newcommand{\eg}{\emph{e.g}. }
\newcommand{\etc}{\emph{etc}. }

\usepackage[pdftex]{graphicx}

\usepackage{enumitem}
\usepackage{wrapfig}
\newcommand{\loss}{\mathcal{L}}
\newcommand{\losswspace}{\loss_{\mathcal{W_+}}}
\newcommand{\lossclip}{\loss_{CLIP}}

\newcommand{\latentmapper}{\mathcal{M}_z}
\newcommand{\generator}{\mathcal{G}}

\newcommand{\rasterizer}{\mathcal{R}}
\newcommand{\textmapper}{\mathcal{M}_t}
\newcommand{\clipmodule}{\mathcal{E}}

\newcommand{\wspace}{\mathcal{W_+}}

\newcommand{\timestep}{t}
\newcommand{\textprompt}{\tau}
\newcommand{\noise}{\epsilon}
\newcommand{\edge}{E}
    \newcommand{\albedoreg}{\mathcal{J}_{a}}
\newcommand{\normalreg}{\mathcal{J}_{n}}
\newcommand{\reg}{\mathcal{J}}

\newcommand{\ourmethod}{PBRGAN\xspace}

\newcommand{\figref}[1]{Fig.~\ref{#1}}
\newcommand{\tabref}[1]{Tab.~\ref{#1}}
\newcommand{\secref}[1]{Sec.~\ref{#1}}

\title{Text-Driven Diverse Facial Texture Generation \\
via Progressive Latent-Space Refinement}

\author {
    Chi Wang\textsuperscript{\rm 12},
    Junming Huang\textsuperscript{\rm 12},
    Rong Zhang\textsuperscript{\rm 3},
    Qi Wang\textsuperscript{\rm 1},
    Haotian Yang\textsuperscript{\rm 2}, \\
    {\bf Haibin Huang\textsuperscript{\rm 2},
    Chongyang Ma\textsuperscript{\rm 2},
    Weiwei Xu\textsuperscript{\rm 1}} \\
    \textsuperscript{\rm 1} State~Key~Lab~of~CAD\&CG,~Zhejiang~University \\ \textsuperscript{\rm 2} Kuaishou Technology \quad \textsuperscript{\rm 3} Zhejiang Gongshang University \\
}

\begin{document}

\maketitle

\begin{figure}[h]
    \centering
    \includegraphics[width=\textwidth]{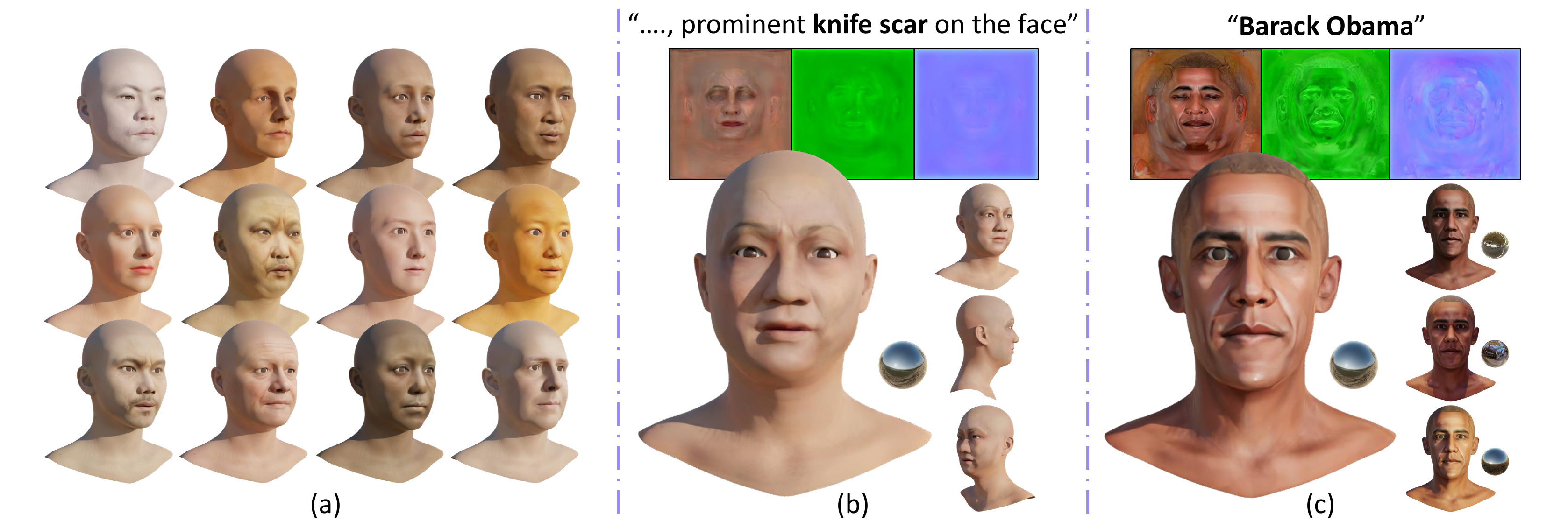}
    \caption{Our method can faithfully generate a variety of facial textures from text prompts for photo-realistic rendering. From left to right: (a) unconditional generation results, (b) multi-view rendering results using our generated PBR texture from an uncommon prompt ``scar'', (c) relighting results using our generated PBR texture of ``Barack Obama''.}
    \label{fig:teaser}
\end{figure}%
\begin{abstract}
Automatic 3D facial texture generation has gained significant interest recently. Existing approaches may 
not support the traditional physically based rendering pipeline or rely on 3D data captured by Light Stage. Our key contribution is a progressive latent space refinement approach that can bootstrap from 3D Morphable Models~(3DMMs)-based texture maps generated from facial images to generate high-quality and diverse PBR textures, including albedo, normal, and roughness. It starts with enhancing Generative Adversarial Networks (GANs) for text-guided and diverse texture generation. To this end, we design a self-supervised paradigm to overcome the reliance on ground truth 3D textures and train the generative model with only entangled texture maps. Besides, we foster mutual enhancement between GANs and Score Distillation Sampling (SDS). SDS boosts GANs with more generative modes, while GANs promote more efficient optimization of SDS. Furthermore, we introduce an edge-aware SDS for multi-view consistent facial structure. Experiments demonstrate that our method outperforms existing 3D texture generation methods regarding photo-realistic quality, diversity, and efficiency. 
\end{abstract}

%
 
\section{Introduction}
\label{sec:intro}

3D facial texture generation has gained widespread attention recently for its potential to boost industries, including Augmented Reality~(AR), Virtual Reality~(VR), gaming, and films. The traditional pipeline of creating textures typically involves UV mapping, material production, or texture synthesis.
Recently, with the advancement of large-scale pre-trained models, such as Stable Diffusion~\cite{rombach2022high} and the Contrastive Language-Image Pre-Training~(CLIP) model, textures can be generated through a more flexible and efficient pathway,~\ie, text-driven texture generation. Pioneers have made effective attempts to obtain facial textures under the guidance of text prompts~\cite{aneja2023clipface, hong_avatarclip_2022}. However, these methods encounter challenges in generating disentangled UV texture maps, such as albedo, roughness, and normal, which are essential to physically based realistic rendering.


This paper focuses on text-driven physically based rendering~(PBR) texture generation to achieve flexible and high-fidelity 3D face creation.
Previous researches employ CLIP loss~\cite{mohammad2022clip, lei2022tango} or Score Distillation Sampling~(SDS)~\cite{poole2022dreamfusion, chen2023fantasia3d} to generate PBR textures with the guidance of given text prompts. Nevertheless, they require extra efforts to retrain the generation model for different 3D objects or text prompts.
DreamFace~\cite{zhang2023dreamface} proposes a diffusion-based generation model in both the latent space and image space, yielding high-quality diffuse texture maps. This method is supervised with explicit ground truth PBR data captured by multi-view photometric systems similar to Light Stage~\cite{ma2007rapid}. The data collection process requires significant resource allocation.
Recently, several approaches~\cite{bai2023ffhquv} utilize 3D Morphable Models~(3DMMs)~\cite{chai2022realy, deng2019accurate} to conveniently synthesize UV textures from large-scale facial image datasets~\cite{karras2020analyzing}, which simplifies the data collection.
However, it is unclear how to seamlessly integrate these scalable datasets into PBR textures generation.

\begin{wrapfigure}{r}{0.49\textwidth}
  \centering
  \includegraphics[width=0.47\columnwidth]{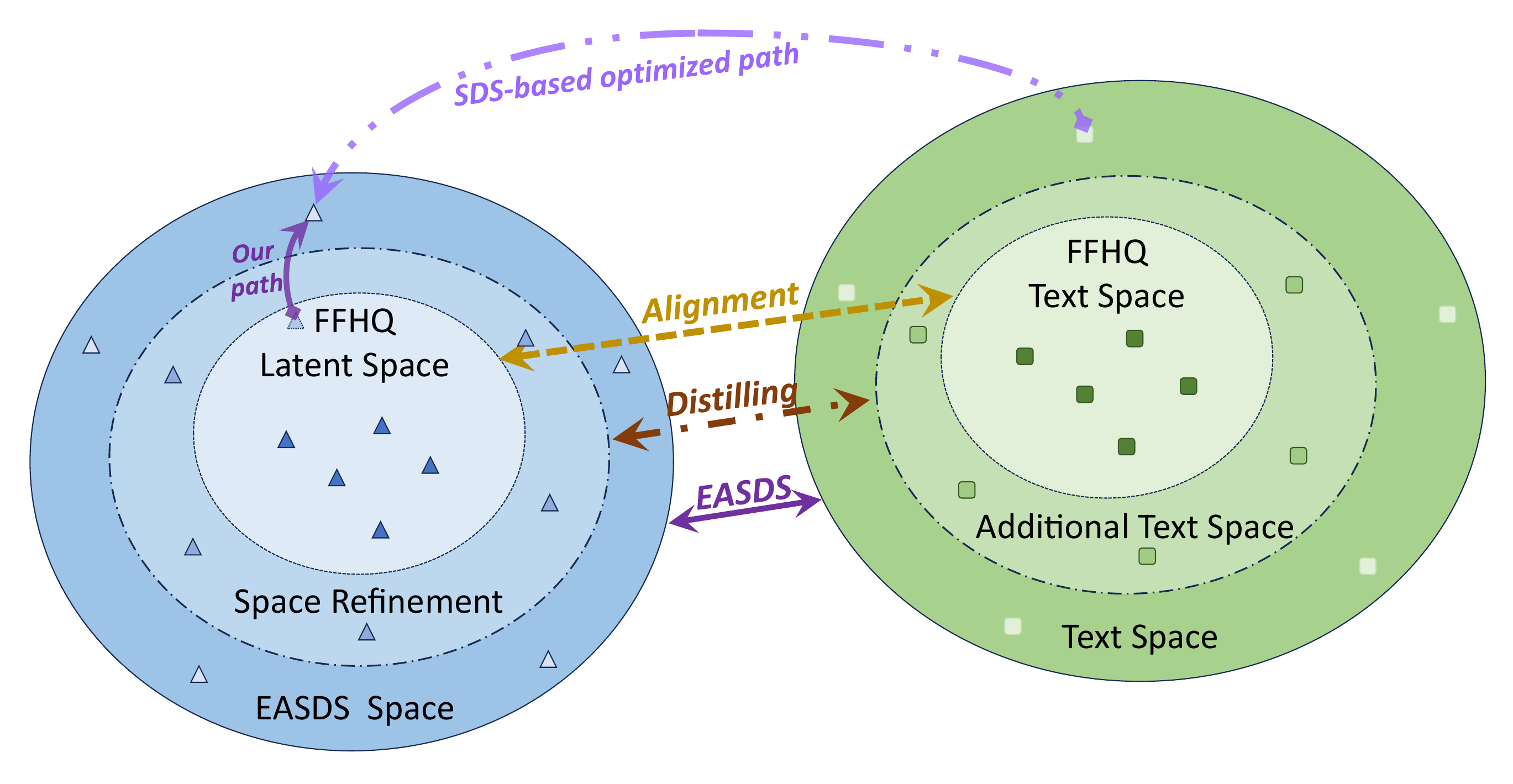}
  \caption{The key idea of latent space refinement approach. The latent space is expanded progressively to handle more text prompts.}
  \label{fig:idea}
\end{wrapfigure}

In this work, we propose a progressive latent space refinement approach that can bootstrap from 3DMM-based texture maps to generate high-quality and diverse PBR textures.
Our key idea is demonstrated in~\figref{fig:idea}.
Our approach starts with initializing a base latent space using the FFHQ-UV dataset~\cite{bai2023ffhquv}. 
Since FFHQ-UV is entangled with weak specular highlights, directly applying them in the generative model training may introduce obvious artifacts in PBR textures, such as light spots. To address this problem, we intend to design a generative network combined with differentiable rendering~\cite{munkberg2022extracting} and well-designed regularization terms to disentangle the textures. Specifically, we employ a GAN model to establish the initial latent space. Once the model is capable of generating reasonable results, we proceed with latent space refinement for text-guided generation and introduce an improved edge-aware SDS to enhance the diversity. 
In the refinement step, the GAN-generated results serve as initializations for SDS, promoting mutual benefits between GAN and SDS. This choice is also due to that it is unclear how to effectively integrate differential rendering with the noisy texture output during the training of denoising network in the diffusion model.  


We construct a framework called \ourmethod with three stages to achieve text-driven, diverse facial PBR texture generation.
(1)~The first stage introduces a self-supervised training scheme to initialize a base latent space for disentangled PBR texture generation. We propose regularization terms regarding material properties, \eg albedo variance or normal deviation, to suppress the specular artifacts.  
(2)~The second stage boosts the latent space with cross-modal generation capability via CLIP-based alignment. It enables our model to generate reasonable results for text attributes resembling the FFHQ distribution.
(3)~In the last stage, we design a bootstrapping framework to distill relevant knowledge of the pre-trained diffusion model to refine the latent space. It leverages the strengths of both GAN and SDS. SDS guides GANs with more generation patterns, while GANs provide better initial positions for SDS. 
We further strengthen the SDS loss with explicit facial feature-line prior and introduce an edge-aware SDS~(EASDS) loss based on ontrolNet~\cite{zhang2023adding}. EASDS helps to align key facial features and effectively improve structural accuracy. After the three stages, PBRGAN can handle uncommon text conditions, such as a knife scar or purple lips, and permits texture generation in a single feedforward process.
In summary, our contributions are as follows:
\begin{itemize}
    \item We propose a bootstrapping approach to refine the latent space for text-driven facial PBR texture generation. It introduces a progressive knowledge-distilling process to handle text conditions by leveraging the mutual benefits of GANs and SDS.
    \item We present a three-stage framework to expand the generative capability. In addition, we design a novel ControlNet-based edge-aware SDS loss to align the facial structures.
    \item Our framework can generate high-quality facial PBR texture maps. Given a specific text prompt, the response time of our method is lower than state-of-the-art (SOTA) methods based on diffusion models.
\end{itemize}

\section{Related Work}
\paragraph{3D-aware image synthesis.}
Generative models have been developed rapidly in recent years~\cite{goodfellow2020generative,esser2021taming,ho2020denoising,rombach2022high}. GAN-based methods, especially StyleGAN~\cite{karras2019style,karras2020analyzing,karras2021alias}, have demonstrated powerful capabilities to generate high-fidelity 2D images. Building upon this, some approaches explored integrating 3D information into image generation to achieve 3D-aware image synthesis.
Visual object network~\cite{NEURIPS2018_92cc2275} proposes a fully differentiable 3D-aware generative model for image and shape synthesis with a disentangled 3D representation. 
Liao~\etal~\cite{liao2020towards} consider the generative process into a 3D content creation stage and a 2D rendering stage, yielding a 3D controllable image synthesis model.
However, 3D data collections are indispensable for these methods. Follow-up works introduce neural representations and leverage 2D data to address the dataset limitation~\cite{nguyen2019hologan, shi2021lifting, xu20223d, deng2022gram, schwarz2022voxgraf}. Pi-GAN~\cite{chan2021pi} and ShadeGAN~\cite{pan2021shading} represent scenes as view-consistent neural radiance fields and map 3D coordinates to pixel values as a 3D prior. EG3D~\cite{chan2022efficient} and Next3D~\cite{sun2023next3d} adopt feature triplanes for more efficient 3D representation. Another series of approaches is dedicated to improving the controllability of 3D-aware generative models~\cite{liu_3d-fm_2022, deng_3d-aware_2023, tan2022volux}. Nonetheless, these methods only generate 2D images and are difficult to use in the classic 3D content creation pipeline. 
\vspace{-10pt}





\begin{figure*}[ht]
  \centering
  \includegraphics[width=\linewidth]{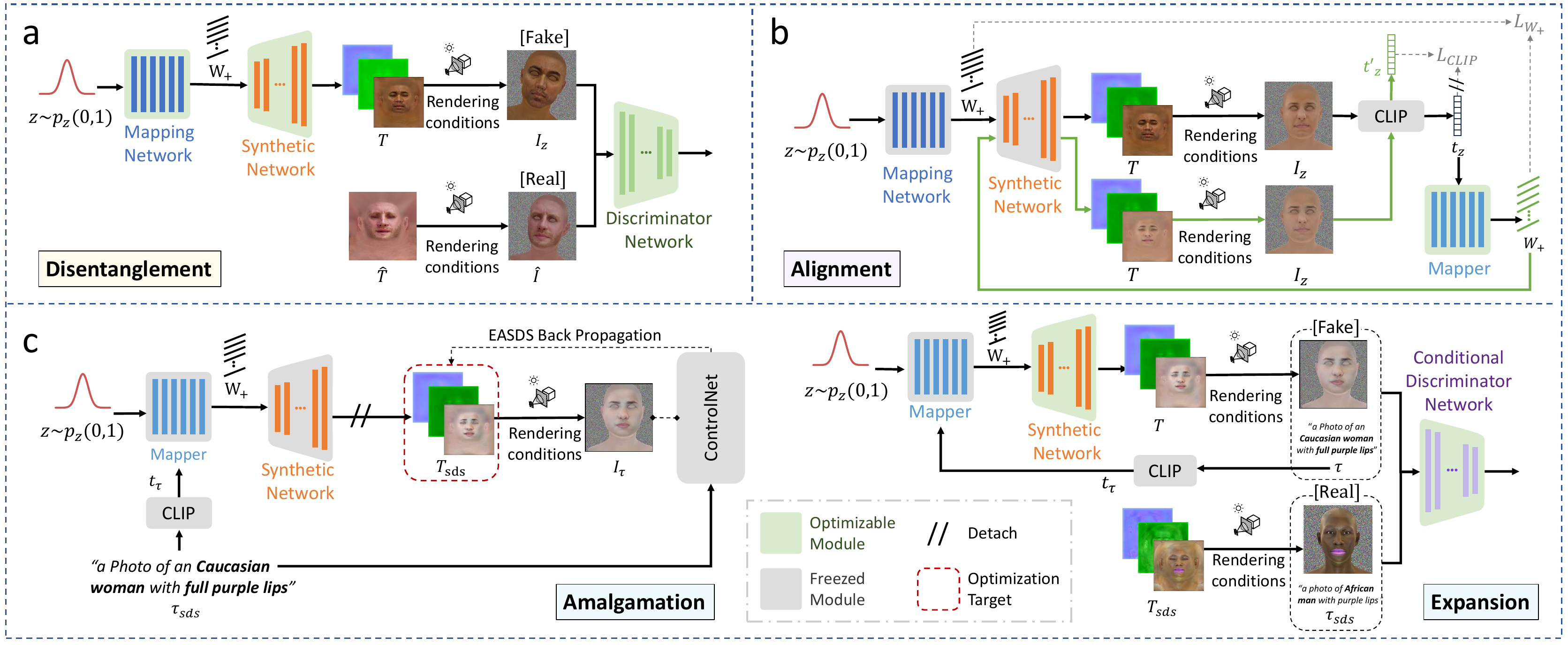}
  \caption{The pipeline of PBRGAN. (a) We generate disentangled PBR textures by leveraging entangled FFHQ-UV textures and differentiable rendering. (b) We align the latent space with the text space under the guidance of CLIP to achieve text-guided generation. (c) We amalgamate GAN and SDS and further expand the latent space to handle more text prompts. }
  \label{fig:pipeline}
\end{figure*}

\paragraph{Text-to-3D generation.}
Generating 3D contents from 2D images has made rapid progress~\cite{gecer2019ganfit, oechsle2019texture, schwarz2020graf, gao2021tm, shen2021deep, gao2022get3d, siddiqui2022texturify} in recent years. But its controllability needs to be further improved.
With the advent of large-scale cross-modal pre-training models, the text-guided 3D generative model~\cite{sanghi2022clip, hong_avatarclip_2022, Jain_2022_CVPR, aneja2023clipface, gecer2019ganfit} has become a frontier research topic. Many researchers utilize the CLIP model~\cite{radford2021learning} to ensure the generated 3D contents match the text prompts. Text2Mesh~\cite{michel2022text2mesh} optimizes the color attributes and geometric displacements of individual vertices simultaneously. 
CLIP-NeRF~\cite{wang2022clip} presents a Nerf-based framework to provide users with flexible control over 3D content using either a text prompt or an exemplar image. However, these methods may fail to generate high-quality results. 

Recently, the powerful generative capabilities of diffusion models have been integrated into text-to-3D generation\cite{wang2023score, metzer2023latent, lin2023magic3d, tang2023dreamgaussian}. 
DreamFusion~\cite{poole2022dreamfusion} adopts pre-trained text-to-image diffusion models as priors and generates realistic 3D models via SDS optimization. ProlificDreamer~\cite{wang2023prolificdreamer} further extends SDS to Variational Score Distillation, which optimizes the distribution of 3D scenes. TEXTure~\cite{richardson2023texture} iteratively renders the object from different viewpoints and applies a depth-to-image 2D diffusion model to generate a complete unwrapped UV texture map. However, the above methods generate textures in entangled representations, which are incompatible with the PBR pipeline.


\vspace{-10pt}
\paragraph{PBR texture generation.}
To fit the existing computer graphics production pipeline, PBR texture generation is essential for photorealistic results. 
Some methods~\cite{yamaguchi2018high, li2020learning, chen2019photo, lattas2021avatarme++} employ 3DMM~\cite{blanz2023morphable, garrido2016reconstruction, tran2018nonlinear, gecer2019ganfit} to generate physical-based texture maps with the guidance of reference images or random noises.
In recent contributions, CLIP loss and SDS loss have been applied to text-driven PBR texture generation, such as CLIP-mesh~\cite{mohammad2022clip}, TANGO~\cite{lei2022tango}, and Fantasia3D~\cite{chen2023fantasia3d}. However, these models must be retrained for each 3D object or each text prompt, which is time-consuming.

DreamFace~\cite{zhang2023dreamface} presents a diffusion model-based scheme to generate personalized 3D faces with text guidance. It can achieve physically-based rendering by decomposing compact latent space into diffuse albedo, specular intensity, and normal maps. The decomposing model relies on a captured high-quality PBR texture dataset, which is usually collected for commercial purposes and has not been released to the public. Besides, the inference efficiency of DreamFace is limited by the diffusion process. In this paper, we expand the $\wspace$ feature space of StyleGAN to support text-to-3D PBR texture generation with no captured datasets. It will benefit from both the efficiency of GANs and the controllability of text-based generation. 


\section{Our Method}

In this section, we present \ourmethod, a progressive latent space refinement approach for text-driven facial texture generation with PBR compatibility and efficient inference. Given the geometry of a 3D face, \ourmethod aims to generate the appearance under the guidance of text prompts. The appearance is decomposed into three components, \ie, albedo, normal, and roughness maps in the texture space without any supervision on explicitly disentangled data.
As illustrated in~\figref{fig:pipeline}, the whole pipeline comprises three stages.
We start by training a PBR StyleGAN~(\secref{sec:first_stage}) to initialize the \ourmethod latent space. 
Then we align it with the text space using a CLIP-based loss~(\secref{sec:second_stage}). Finally, we further refine the \ourmethod latent space for more uncommon text conditions~(\secref{sec:third_stage}).

\subsection{PBR StyleGAN}
\label{sec:first_stage}

Given an arbitrary 3D facial geometry, we propose a self-supervised learning scheme to generate disentangled UV textures,~\eg, the diffuse/specular/normal maps, via leveraging a non-PBR facial texture dataset FFHQ-UV~\cite{bai2023ffhquv}. 
FFHQ-UV is constructed with even illumination conditions. Its textures are entangled with weak specular highlights. Directly applying them to train the generation model may introduce light spot artifacts. To avoid this issue, we do not utilize FFHQ-UV data as supervision. Instead, we train a disentangling model with differentiable rendering and additional regularization terms. 

As diffusion models typically involve iterative noise injection and denoising processes, the changes in the image are subtle at each iteration. The non-deterministic and complex nature of its iterative gradient computation process presents challenges for integrating diffusion models with differentiable rendering during the denoising process. It hinders us from directly training a diffusion model with FFHQ-UV output. Therefore, we construct a PBR texture generator based on StyleGAN2~\cite{karras2020analyzing} to establish an initial latent space. 
The following sections will provide detailed introductions to the network architecture and training strategy.

\vspace{-10pt}
\paragraph{Network structure.}
\figref{fig:pipeline}(a) demonstrates the first stage of our \ourmethod framework.
Given a randomly initialized latent code $z\in\mathbb{R}^{512}$, a latent mapper $\latentmapper$ transforms $z$ into an intermediate space $\wspace$ to produce a new latent embedding $w\in\mathbb{R}^{16*512}$. Then $w$ modulates the parameters of $\generator$ through adaptive instance normalization (AdaIN)~\cite{huang2017arbitrary}. 
To enable StyleGAN with physically-based rendering, a generator $\generator$ outputs a seven-channel appearance map in the unwrapped UV space. Following the setting of Munkberg~\etal~\cite{munkberg2022extracting}, the output can be split into $T=(T_a, T_r, N)$, where $T_a$ is the albedo map, $T_r$ are roughness parameters, and $N$ describes tangent space normal perturbations.
For the geometry, we generate random facial meshes by sampling in the parameter space of the 3DMM-based model HIFI3D++~\cite{chai2022realy}. It can provide abundant geometric information with enhanced diversity.

\vspace{-10pt}
\paragraph{Differentiable-rendering-based training.}
We introduce a differentiable rendering module $\rasterizer$ to train the \ourmethod.
$\rasterizer$ takes the predicted PBR textures, a 3D geometry $s$, and a random pose $p$ as input and produces a 2D facial image $I_z=\rasterizer(T_a, T_r, N, s, p)$. For the ground truth, $\rasterizer$ utilizes an entangled texture map $\hat{T}$ in FFHQ-UV to render another image $\hat{I}=\rasterizer(\hat{T}, s, p)$. 
Then \ourmethod can be supervised by the distance between $I_z$ and $\hat{I}$.
Besides, $I_z$ and $\hat{I}$ are input to a global discriminator and a patch-based local discriminator for detail enhancement. We employ the discriminator training technique of Diffusion-GAN~\cite{wang2022diffusion} to improve the diversity of the results. 

\vspace{-10pt}
\paragraph{Regularization terms.}
Moreover, we design two regularization terms to stabilize the training. They introduce priors of material properties to guide the texture disentanglement. 
Specifically, we propose $\albedoreg$ for the albedo and $\normalreg$ for the normal map. $\albedoreg$ is used to reduce the specular highlights by constraining the albedo of the skin region to be smooth. $\normalreg$ constrains the predicted normals should stick to a standard surface normal of the UV space.
The regularization term $\reg$ can be represented as:
\begin{equation}
    \begin{aligned}
        \reg &= \albedoreg + \normalreg \\
             &= | \nabla (T_a \odot M_{skin})| +| T_n - N| 
    \end{aligned}
\end{equation}
where $M_{skin}$ is the mask of the skin region, $\odot$ is elment-wise multiplication, $\nabla$ represents the total variation, and $N$ is a standard normal map with $[0,0,1]$ for all pixels. $\albedoreg$ is calculated in both the RGB space and the hue channel of the HLS color space.

\subsection{Text and Latent Alignment}
\label{sec:second_stage}

With the emergence of pre-trained cross-modal models, 
text prompts can provide a flexible and user-friendly interface for 3D content generation. Existing approaches typically rely on fine-tuning the generative models to establish consistency between the results and the texts\cite{poole2022dreamfusion}. However, it usually requires a lot of time for optimization. To avoid this issue, we extend StyleGAN to support text-based PBR texture generation. The key idea is correlating the text embeddings with the $\wspace$ latent space of \ourmethod. 

Recent advances based on GAN inversion techniques and cross-modal models have led to a more feasible solution for aligning text with latent space.
GAN-inversion-based image editing algorithms~\cite{zhu2020domain, abdal2020image2stylegan++, shen2021closed} invert given images back into the latent space of GANs and manipulate them by varying the latent code in different interpretable directions, such as age and expression. 
In the meantime, the CLIP model has made remarkable progress in establishing cross-modal correlations between image and text space~\cite{pinkneyclip2latent}. Therefore, we employ the CLIP model as an anchor point for alignment between text space and $\wspace$ space. We will introduce the text prompt generation and the pipeline in the following.

\vspace{-10pt}
\paragraph{Text prompt generation.} We define a text prompt dictionary with geometry-independent facial attributes to construct the text space. The dictionary consists of common facial attributes, such as ``eyebrow shape'' or ``red lips'', and uncommon attributes like ``scar'' or ``purple lips''. We formulate different text prompts through randomly sampling attributes from the dictionary.
For more details, please refer to our supplementary materials.

\vspace{-10pt}
\paragraph{The alignment framework.}
As shown in~\figref{fig:pipeline}(b), we integrate an additional text embedding module into the \ourmethod to achieve the alignment.
It consists of a CLIP embedding module $\clipmodule$ and a text mapper $\textmapper$ with a twining structure of the latent mapper $\latentmapper$ in~\secref{sec:first_stage}. 
In this stage, the parameters of $\generator$ and $\clipmodule$ are frozen. $\textmapper$ is optimized to map the text embedding space $\mathcal{W'_+}$ to $\wspace$. To train $\textmapper$, we random sample a latent $z$ and produce an image $I_z=\rasterizer(\generator(z), s, p)$. Then $I_z$ is fed into $\clipmodule$ as input to get the embedding $t_z$ in the CLIP space. The text mapper $\textmapper$ takes $t_z$ as input and transforms it into $w'\in\mathcal{W'_+}$. $w'$ is utilized to generate a facial image, which can be represented as:
\begin{equation}
\label{eq:text_mapping}
    I_t=\rasterizer(\generator(\textmapper(\clipmodule(I_z))), s, p).
\end{equation}

\vspace{-10pt}
\paragraph{The alignment loss functions.}
The space alignment is supervised by two loss functions, \ie, a loss $\losswspace$ in the $\wspace$ space and a loss $\lossclip$ in the CLIP space. $\losswspace$ is the Manhattan distance between $\mathcal{W'_+}$ and $\wspace$. $\lossclip$ is a cycle consistency loss to constrain the rendered image $I_t$ to have a consistent CLIP embedding with $t_z$. The overall loss function is as follows:
\begin{equation}
\label{eq:second_loss}
    \loss=|\textmapper(\clipmodule(I_t)) - \latentmapper(z)|+\alpha|\clipmodule(I_z)-\clipmodule(I_t)|,
\end{equation}
where $\alpha$ is the weight for $\lossclip$.
In our experiments, $\alpha$ is set to $0.05$.
Once the training of the alignment module is done, we can generate PBR textures for a given text prompt $t$ through a feedforward pass $\generator(t)$.

\subsection{Latent Space Refinement}\label{sec:third_stage}

After the alignment, \ourmethod is proficient at generating images from common text prompts resembling the FFHQ distribution. It still faces limitations in producing reasonable textures using uncommon prompts, \eg, ``a prominent knife scar on the face''.
To generate textures that match these prompts, existing methods~\cite{wang2023score, metzer2023latent} employ a pre-trained diffusion model as an image prior and perform parameters optimization using the SDS loss~\cite{poole2022dreamfusion}. Such approaches require inefficient model optimization for each input prompt.

In the third stage of \ourmethod, we propose a latent space refinement approach to amalgamate GANs and SDS so that they can benefit from each other mutually. SDS can provide more generation pattern guidance for GANs to break through space limitations, while GANs can produce better initial positions for the optimization of SDS to reduce the computational cost.
Meanwhile, we expand the latent space to efficiently generate more diverse PBR textures by leveraging the generative capability of SDS.

\vspace{-10pt}
\paragraph{GANs and SDS amalgamation.}
To amalgamate GANs and SDS, we use results generated by GANs as the optimization target of the SDS loss~(amalgamation part of~\figref{fig:pipeline}(c)). 
Furthermore, we modify the basic SDS loss into an enhanced edge-aware version (EASDS) by introducing a key facial structure alignment module and a pre-trained ControlNet~$\phi$~\cite{zhang2023adding}. The EASDS has four steps (\figref{fig:easds}):

\begin{wrapfigure}{r}{0.5\textwidth}
  \centering
  \includegraphics[width=0.48\columnwidth]{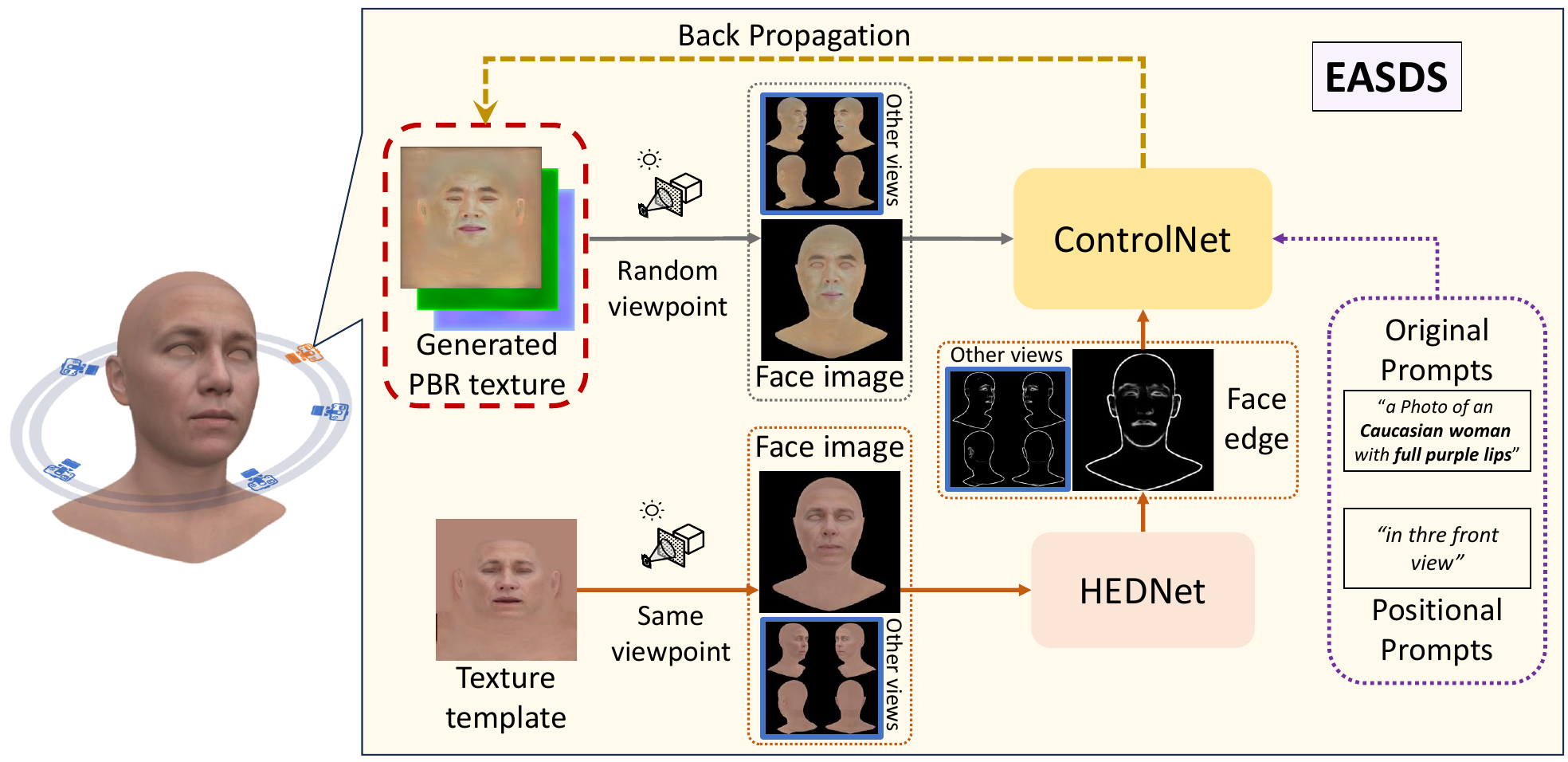}
  \caption{Explanation of our edge-aware SDS (EASDS). We sample from different viewpoints and employ soft edges detected from a template texture as feature-line conditions for ControlNet.}
  \label{fig:easds}
\end{wrapfigure}

\begin{itemize}
    \item[(1)] Optimization target preparation. We run an inference procedure of \ourmethod to generate the texture maps $T$ for a given text prompt $\textprompt$. $T$ will be taken as the target parameter for the optimization of EASDS. 
    \item[(2)] Camera pose setting. We place a series of predefined viewpoints around the face model to fit our differentiable- rendering-based pipeline.
    \item[(3)] Feature-line extraction. We render reference facial images $I_{ref}$ from the
    viewpoints with a random facial geometryand a predefined template texture. Then a pre-trained edge detection network HEDNet~\cite{xie2015holistically} is exploited to detect soft edges $\edge$ from the renerderred image. 
    \item[(4)] EASDS optimization. For a given viewpoint, the rasterizer renders a facial image $I_{\textprompt}=\rasterizer(T,s,p)$ with the same geometry of~(3). The rough HED edges associated with this viewpoint from the former step are taken as feature-line priors to ControlNet. We optimize $T$ with the ControlNet gradient through the rendered image $I_{\textprompt}$. The ControlNet $\phi$ use $\textprompt$ and $\edge$ to provide a score function $\hat{\noise}_\phi$ as the optimization guidance for $T$:
\begin{equation}
\nabla_{T} \mathcal{L}_{\mathrm{EASDS}}(\phi, I_\textprompt) \triangleq \mathbb{E}_{\timestep, \noise}
\left[
    w(\timestep)
    \left(
        \hat{\noise}_\phi 
        \left( 
            \mathbf{z}^{I_\textprompt}_{\timestep}; \textprompt, \timestep, \edge 
        \right)-\noise
    \right) 
    \frac{\partial I_\textprompt}{\partial T}
\right]
\end{equation}

where $w(\timestep)$ is a timestep-related weighting function, $\noise$ is the sample noise, $\mathbf{z}^{I_\textprompt}_{\timestep}$ is the noisy image based on ${I_\textprompt}$.
The optimization starts from the front view and iteratively proceeds to other viewpoints. With feature lines extracted from the same 3D reference model, EASDS ensures that the images rendered from different perspectives maintain multi-view consistency in facial features.
    
\end{itemize}

\begin{figure*}[ht]
    \centering
    \includegraphics[width=\linewidth]{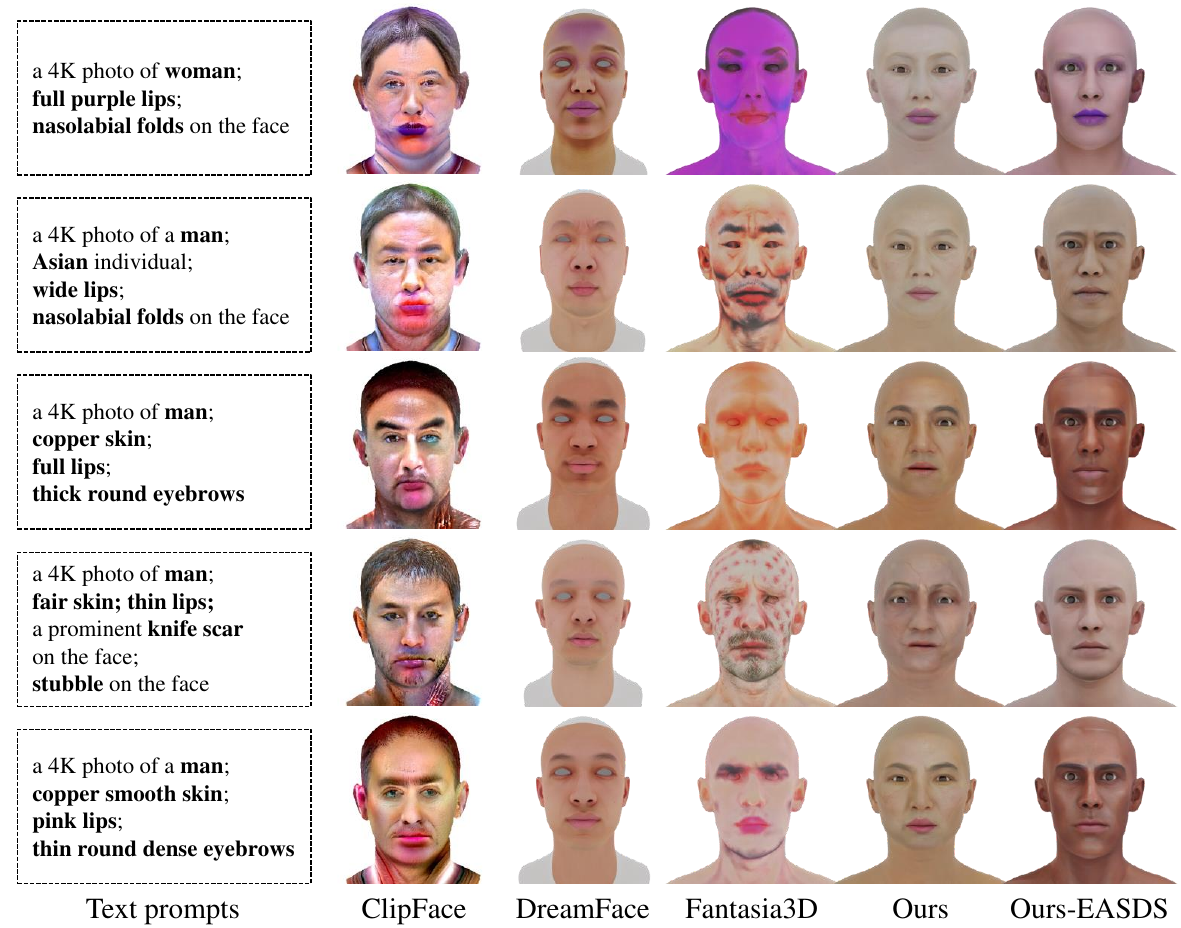}
\caption{Qualitative comparison results. From left to right: the input text prompts, the results generated with ClipFace, DreamFace, Fantasia3D, ours, and ours-EASDS.}
\label{fig:40_vssota}
\vspace{-4mm}
\end{figure*}

For special text prompts of specific individuals, such as Barack Obama, Avatar, Iron Man, \etc, \ourmethod will produce the corresponding texture maps via direct optimization based on EASDS.

\vspace{-10pt}
\paragraph{Latent space expansion.}

For some uncommon attributes that FFHQ-UV does not possess, such as ``scar'' or ``purple lips'', we expect to distill relevant knowledge of the diffusion model and inject it into the latent space of \ourmethod. It will expand the latent space and boost \ourmethod for better generative diversity. Specifically, we first construct two kinds of text prompts. Common prompts ${\tau_{c}}$ only consist of common attributes, while uncommon prompts ${\tau_{uc}}$ involve uncommon attributes. Then, we generate a series of facial PBR textures based on these prompts through EASDS. Afterward, we employ the generated textures as the manufactured ground truth data to finetune the \ourmethod. 


As illustrated in the expansion part of~\figref{fig:pipeline}(c), the finetuning procedure has two training paths: one is an unconditional generation path from random sampled $z$, which can stabilize the training; the other is a conditional generation path based on mixed prompts of ${\tau_{c}}$ and ${\tau_{uc}}$. For the latter one, we introduce a conditional discriminator~\cite{karras2020training} to match the generated results with the input conditions.
The conditional discriminator is initialized with parameters of the global discriminator (see \secref{sec:first_stage}). It takes the rendered images and text embeddings as input and produces true value only when the image is matched with the prompt.
After network training, the generative space of \ourmethod is expanded to conditions beyond the FFHQ distribution.



%

%



\section{Experiments}
\begin{table*}
    \centering
    \resizebox{\linewidth}{!}{
        \begin{tabular}{l|l|l|l|l|c|c|c|c|c}
        \toprule
        & \multirow{2}{*}{Method}   & \multirow{2}{*}{Rep} & \multirow{2}{*}{Data} & \multirow{2}{*}{Inf Time / min} & \multicolumn{4}{c|}{Metrics} & \multirow{2}{*}{User Study} \\  \cline{6-9} 
                    &              &                      &                       &                             & hyperIQA~\cite{Su2020hyperiqa}~$\uparrow$ & CLIPIQA$+$~\cite{wang2022clipiqa}~$\uparrow$ & TOPIQ$_{face}$~\cite{chen2023topiq}~$\uparrow$ & CLIPSCORE~\cite{hessel2021clipscore}~$\uparrow$  \\ 
        \midrule[0.5pt]
        \multirow{2}{*}{w/o prompt}  & ClipFace~\cite{aneja2023clipface}                  & Texture                  & FFHQ                  & $<$ 0.01 (1x 3090)                                                                & 0.63                                      &           0.55                           & 0.58                                               &-    &-    \\
        & Ours.                     & PBR                  & FFHQUV                & $<$ 0.01 (1x3090)                                                                & \first{0.74}                                      &             \first{0.57}                            &  \first{0.64}                                          &-        &- \\ \midrule[0.5pt]
        \multirow{6}{*}{w/ prompt} &TANGO~\cite{lei2022tango} & PBR                 & -                     & 10 (1x 3090)                                                          & 0.50                                      &           0.45                                   &               0.35                                 &0.67   & 1.61    \\
        & Fantasia3D~\cite{chen2023fantasia3d}                & PBR                 & -                     & 45 (8x 3090)                                                             &  0.66                                     &  0.52                                         &     0.57                                           & 0.74    & 2.20     \\
        & ClipFace~\cite{aneja2023clipface}                  & Texture                  & FFHQ                  & 90 (1x 3090)                  & 0.63                                      &           0.56                           & 0.52                                               & \first{0.77}     & 2.36   \\
        & DreamFace~\cite{zhang2023dreamface}                 & PBR                  & LightStage            & 5 (1x A6000)                                  &       \first{0.76}                                    &                       \first{0.63}                       & \first{0.77}    &0.72                                        & \second{2.96}   \\ \cmidrule{2-10} 
        & Ours                    & PBR                  & FFHQUV                & $<$ 0.01 (1x3090)                                                             & \second{0.72}                                      &             \second{0.60}                            &  0.73                                              & \second{0.76}    & 2.80    \\
        & Ours-EASDS                & PBR                  & FFHQUV                & 1 (1x 3090)                                         &          0.68                                 &     0.58             &  \second{0.74}                            &  0.74               & \first{3.34}                        \\ \bottomrule
        \end{tabular}
    }
  \caption{Qualitative comparison results. We conduct experiments with or without prompts. The best results are highlighted as \first{1st} and \second{2nd}.  Our method wins first place in the human perceptual study and inference time. We also achieve the second-best in other metrics for text-conditional generation. Note that DreamFace needs captured PBR data to train and spends five times as much inference time as ours.}
  \vspace{-3mm}
  \label{tab:vssota}
\end{table*}

In this paper, we focus on text-driven facial PBR texture generation. The experiments are conducted on a Linux server with 8 Tesla V100 GPUs and an Intel Xeon Platinum 8268 CPU. We demonstrate details of experiments, comparisons with state-of-the-art methods, and ablation studies.  

\subsection{Experimental Setup}

\paragraph{Dataset preparation.}
We use the FFHQ-UV dataset~\cite{bai2023ffhquv} for experiments. FFHQ-UV is a large-scale 3DMMs-based facial UV-texture dataset derived from a natural image dataset FFHQ. It contains $50,000+$ texture UV-maps with even illuminations and cleaned facial regions. For the facial geometry, We randomly sample $\numprint{10000}$ facial mesh data from HIFI3D++~\cite{chai2022realy}. Since HIFI3D++ contains no eyeballs, we employ the albedo and geometry of the eyeballs produced from the pipeline of FFHQ-UV~\cite{bai2023ffhquv}. To get rid of the influence of hair, we fill the hair area with the average skin color and smooth the border region.


\vspace{-4mm}
\paragraph{Implementation details.}

We train \ourmethod using rendered images of a resolution of $512 \times 512$ and PBR texture maps of the same size.
The detailed settings are as follows:
\begin{itemize}[leftmargin=*]
  \item \textbf{The first stage~(\secref{sec:first_stage}):} learning rate = $0.0025$, batch size = $8$. We use NVDIFFREC~\cite{munkberg2022extracting} as our differentiable rendering module.
  \item \textbf{The second stage~(\secref{sec:second_stage}):} learning rate = $1$, batch size = $32$. For CLIP supervision, we adopt the `ViT-B/32' pre-trained variant. To get better CLIP embedding of the rendered results, we use 3D geometries with eyeballs here.
  \item \textbf{The third stage~(\secref{sec:third_stage}):}
  learning rate = $0.0025$, batch size = $8$. 
  For EASDS supervision, we utilize the HED Boundary-conditional ControlNet~\cite{zhang2023adding} with the base model Stable Diffusion v1.5~\cite{rombach2022high}. We predefine $20$ viewpoints for the EASDS optimization.
  To feed viewpoint information into the EASDS optimization, a viewpoint-relevant positional prompt, structured as ``in the front/back/left/right/bottom/vertical view'', is appended to the original prompt. In addition, We remove the face-related prompts, such as lips or eyebrows, when optimized in the back view since SDS tends to generate human faces on the back of the head in order to fit these prompts. 
\end{itemize}

\subsection{Comparison and Analysis}

We compare our method with four state-of-the-art texture generation methods, including Tango~\cite{lei2022tango}, Fantasia3D~\cite{chen2023fantasia3d}, ClipFace~\cite{aneja2023clipface}, and DreamFace~\cite{zhang2023dreamface}.
ClipFace can be further divided into two parts: unconditional sampling and conditional optimization, while all the other methods only support conditional generation.
Therefore, we first compare our method with ClipFace on the unconditional generation task and then make a comparison with all methods on the text-conditional generation task.

\vspace{-2mm}
\paragraph{Metrics.}
We use the following metrics\footnote{We use an implementation of IQA provided by~\cite{pyiqa}.} from three perspectives to evaluate our method.
\begin{itemize}
\item \textbf{Non-reference image quality assessment~(IQA) metrics:} Non-reference IQA aims to evaluate the quality of images without relying on a reference image. We adopt MUSIQ~\cite{ke2021musiq}, HyperIQA~\cite{Su2020hyperiqa}, and CLIPIQA+~\cite{wang2022clipiqa} for the generated image quality evaluation. 
\item \textbf{Task-specific metrics:} We utilize a tailored metric for face quality measurement, \ie TOPIQ$_{face}$~\cite{chen2023topiq}, which is trained with face IQA dataset GFIQA~\cite{gfiqa20k}, to evaluate the quality of the rendered facial images.
\item \textbf{Prompt-aware metrics:} We use the CLIPSCORE~\cite{hessel2021clipscore} to measure the alignment between the given prompt and the corresponding generated face. 
\end{itemize}
For these metrics, higher scores indicate better image quality or higher text-image consistency.

\vspace{-2mm}
\paragraph{Unconditional texture generation.}
For the text-unrelated generation task, we use non-reference-IQA and task-specific metrics to evaluate our method with $\numprint{1000}$ randomly sampled latent code $z$.
As shown in the ``without prompt'' part of Table~\ref{tab:vssota}, our method achieves a higher score by $10.3\%$ on TOPIQ$_{face}$, which shows the effectiveness of our \ourmethod. Our method also outperforms ClipFace in terms of the non-reference-IQA metrics, yielding better image quality with enhanced sharpness and intricate details. It demonstrates the effectiveness of our latent space representation.

\vspace{-2mm}
\paragraph{Text-driven texture generation.}
For the conditional generation task, our methods were evaluated by randomly sampling $30$ prompts from our dictionary to cover different attributes. Specifically, we first synthesize the images with these prompts using all methods. The images are combined with corresponding text prompts to generate text-image pairs. Then, we adopt all the above metrics to evaluate the image quality and text-image consistency of these pairs. 

Since \ourmethod is capable of generating PBR texture maps with straightforward inferences, our method strikes a balance between efficiency and performance. As illustrated in the ``with prompt'' part of \tabref{tab:vssota}, \ourmethod based on EASDS
 optimization~(Ours-EASDS) is able to generate textures at $1$- minute level. It shows that the amalgamation of GAN and SDS can reduce the optimization time by a large margin. For the quality evaluation, Ours-EASDS achieves comparable results to DreamFace with tenuous distinction and outperforms all other methods. Note that DreamFace is trained with captured PBR texture maps, while our method involves no ground truth PBR data. Further expanding the latent space with EASDS-based textures~(Ours) may slightly diminish partial quantitative metrics, but it decreases the inference time to less than $0.5$ seconds.

 \figref{fig:40_vssota} demonstrates the qualitative comparison results. Although ClipFace can generate facial textures with most designated attributes, its results have obvious artifacts~(2nd Column). Fantasia3D is designed for generic-object texture generation. Applying it to facial texture generation may cause disharmonious patterns~(4th Column). DreamFace is able to generate relatively realistic results. However, it fails to handle uncommon attributes like ``scar'' or ``purple lips''~(3rd Column). Our methods are capable of producing reasonable facial texture maps with diverse prompts. 

\vspace{-1em}
 \paragraph{User study.}
We conducted a user study to evaluate the perceptual quality of the results. 
We generate $30$ text-image pairs for each method of TANGO, Fantasia3D, ClipFace, DreamFace, OurS and Ours-EASDS. $150$ participants are recruited to evaluate the results. 
We ask every participant to grade each text-image pair according to the image quality and text-image consistency. The grading is quantified as a preference score to assess the overall quality from 1~(least low quality) to 5~(most high quality). As the last column of~\tabref{tab:vssota} shows, ours-EASDS is superior to other methods.

\subsection{Ablation Study}
To explore the effects of our modules, we conducted ablation experiments for EASDS and GAN-SDS amalgamation.

\vspace{-3mm}
\paragraph{EASDS.}
\begin{wrapfigure}{r}{0.5\textwidth}
    \centering
    \includegraphics[width=0.48\textwidth]{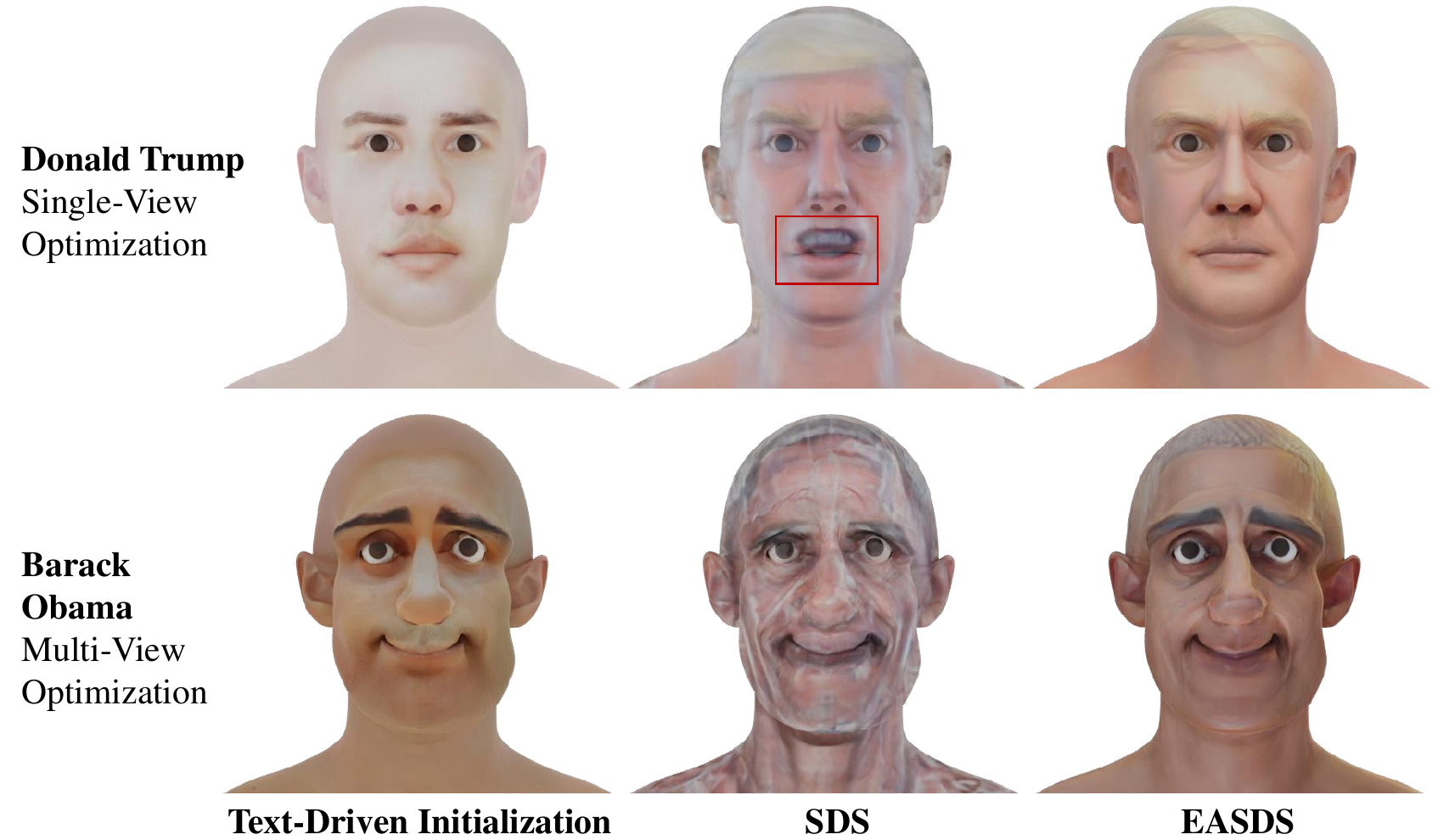}
    \caption{Ablation study results of EASDS. Note that there are artifacts and misalignments in the results of SDS. The second row shows the generated textures using challenging geometry under the guidance of ``Barack Obama''.  Benefiting from EASDS's alignment of feature lines, our method is capable of generating better textures tailored for extreme geometries.}
    \label{fig:41_as_sds}
    \vspace{-1em}
\end{wrapfigure}

We compare our EASDS with the classical SDS~\cite{poole2022dreamfusion}, which has no edge priors. Specifically, given a text prompt, we generate an initial PBR texture with our \ourmethod of the second stage. Afterward, we perform SDS-based and EASDS-based optimization separately to generate different texture maps. As the \figref{fig:41_as_sds} shows, there are some misalignments between the geometry edges and texture edges in the SDS-based results. With additional feature-line priors, our EASDS can enhance facial structural accuracy and generate better PBR textures than SDS.

\begin{wrapfigure}{r}{0.5\textwidth}
    \centering
    \includegraphics[width=0.48\textwidth]{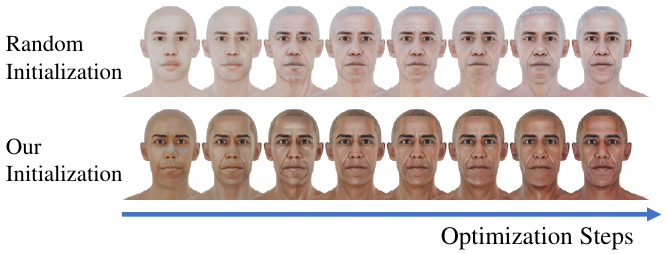}
\caption{The influence of the EASDS initialization.}
\vspace{-4mm}
\label{fig:40_easds_init}
\end{wrapfigure}

\vspace{-3mm}
\paragraph{GAN and SDS amalgamation.}
The classical SDS-based texture generation methods \eg Fantasia3D~\cite{chen2023fantasia3d} are optimized from randomly initialized parameters. The initial optimization point is distant from the target. In this paper, we utilize the results generated by \ourmethod of the second stage as the initial point for EASDS, which is beneficial for shortening the optimization path and generating better results (\figref{fig:40_easds_init}). The amalgamation can reduce the required optimization time for each prompt from $70$ to $20$ seconds.

\section{Conclusion}
In this work, we present a progressive latent space refinement approach to generate high-quality and diverse facial PBR textures under the guidance of text prompts.
Our approach constructs a multi-stage framework to bootstrap from 3DMM-based textures to reduce the reliance on ground truth PBR data.
We design a self-supervised scheme with differentiable rendering and carefully crafted regularization terms to disentangle UV textures.
To refine the latent space, we introduce a knowledge distillation process utilizing GANs and SDS.
We also propose a ControlNet-based and edge-aware SDS to increase the accuracy of facial structures.
The experiments demonstrate that our method can efficiently create high-quality and diverse facial PBR textures.
In the future, we aim to explore the generation of accurate facial geometry in conjunction with diverse PBR textures.

\medskip


{
    \small
    \bibliographystyle{plain}
    \bibliography{nips.bib}
}

%
%
%
%


\end{document}